# Deep Learning for Detecting Building Defects Using Convolutional Neural Networks

Husein Perez, Joseph H. M. Tah and Amir Mosavi

Oxford Institute for Sustainable Development, School of the Built Environment, Oxford Brookes University, Oxford OX3 0BP, UK; a.mosavi@brookes.ac.uk

**Abstract:** Clients are increasingly looking for fast and effective means to quickly and frequently survey and communicate the condition of their buildings so that essential repairs and maintenance work can be done in a proactive and timely manner before it becomes too dangerous and expensive. Traditional methods for this type of work commonly comprise of engaging building surveyors to undertake a condition assessment which involves a lengthy site inspection to produce a systematic recording of the physical condition of the building elements, including cost estimates of immediate and projected long-term costs of renewal, repair and maintenance of the building. Current asset condition assessment procedures are extensively time consuming, laborious, and expensive and pose health and safety threats to surveyors, particularly at height and roof levels which are difficult to access. This paper aims at evaluating the application of convolutional neural networks (CNN) towards an automated detection and localisation of key building defects, e.g., mould, deterioration, and stain, from images. The proposed model is based on pre-trained CNN classifier of VGG-16 (later compaired with ResNet-50, and Inception models), with class activation mapping (CAM) for object localisation. The challenges and limitations of the model in real-life applications have been identified. The proposed model has proven to be robust and able to accurately detect and localise building defects. The approach is being developed with the potential to scale-up and further advance to support automated detection of defects and deterioration of buildings in real-time using mobile devices and drones.

1. Introduction

Clients with multiple assets are increasingly requiring intimate knowledge of the condition of each of their operational assets to enable them to effectively manage their portfolio and improve business performance. This is being driven by the increasing adverse effects of climate change, demanding legal and regulatory requirements for sustainability, safety and well-being, and increasing competitiveness. Clients are looking for fast and effective means to quickly and frequently survey and communicate the condition of their buildings so that essential maintenance and repairs can be done in a proactive and timely manner before it becomes too dangerous and expensive [1-4]. Traditional methods for this type of work commonly comprise of engaging building surveyors to undertake a condition assessment which involves a lengthy site inspection resulting in a systematic recording of the physical conditions of the building elements with the use of photographs, note taking, drawings and information provided by the client [5-7]. The data collected are then analysed to produce a report that includes a summary of the condition of the building and its elements [8]. This is also used to produce estimates of immediate and projected long-term costs of renewal, repair and maintenance of the building [9, 10]. This enables facility managers to address current operational requirements, while also improving their real estate portfolio renewal forecasting and addressing funding for capital projects [11]. Current asset condition assessment procedures are extensively time consuming, laborious and expensive, and pose health and safety threats to surveyors, particularly at height and roof levels which are difficult to access for inspection [12].

Image analysis techniques for detecting defects have been proposed as an alternative to the manual on-site inspection methods. Whilst the latter is time-consuming and not suitable for quantitative analysis, image analysis-based detection techniques, on the other hand, can be quite challenging and fully dependent on the quality of images taken under different real-world situations (e.g. light, shadow, noise, etc.). In recent years, researchers have experimented with the application of a number of soft computing and machine learning-based detection techniques as an attempt to increase the level of automation of asset condition inspection [14-20]. The notable efforts include; structural health monitoring with Bayesian method [13], surface crack estimation using Gaussian regression, support vector machines (SVM), and neural networks [14], SVM for wall defects recognition [15], crack-detection on concrete surfaces using deep belief networks (DBN) [16], crack detection in oak flooring using ensemble methods of random forests (RF) [17], deterioration assessment using fuzzy logic [18], defect detection of ashlar masonry walls using logistic regression [19-20]. The literature also includes a number of papers devoted to the detection of defects in infrastructural assets such as cracks in road surfaces, bridges, dams, and sewerage pipelines [21–30]. The automated detection of defects in earthquake damaged structures has also received considerable attention amongst researchers in recent years [31–33]. Considering these few studies, far too little attention has been paid to the application of advanced machine learning methods and deep learning methods in the advancements of smart sensors for the building defects detection. There is a general lack of research in the automated condition assessment of buildings; despite they represent a significant financial asset class.

The major objective of this research has therefore is set to investigate the novel application of deep learning method of convolutional neural networks (CNN) in automating the condition assessment of buildings. The focus is to automated detection and localisation of key defects arising from dampness in buildings from images. However, as the first attempt to tackle the problem, this paper applies a number of limitations. Firstly, multiple types of the defects are not considered at once. This means that the images considered by the model belong to only one category. Secondly, only the images with visible defects are considered. Thirdly, consideration of the extreme lighting and orientation, e.g., low lighting, too bright images are not included in this study. In the future works, however, these limitations will be considered to be able to get closer to the concept of a fully automated detection.

The rest of this paper is organized as follows. A brief discussion of a selection of the most common defects that arise from the presence of moisture and dampness in buildings is presented. This is followed by a brief overview of CNN. These are a class of deep learning techniques primarily used for solving fundamental problems in computer vision. This provides the theoretical basis of the work undertaken. We propose a deep learning-based detection and localisation model using transfer learning utilising the VGG-16 model [40] for feature extraction and classification. Next, we briefly discuss the localisation problem and the class activation mapping (CAM) [41] technique which we incorporated with our model for defect localisation. This is followed by a discussion of the dataset used, the model developed, the results obtained, conclusions and future work.

## 2. Dampness in Buildings

Buildings are generally considered durable because of their ability to last hundreds of years. However, those that survive long have been subject to care through continuous repair and maintenance throughout their lives. All building components deteriorate at varying rates and degrees depending on the design, materials and methods of construction, quality of workmanship, environmental conditions and the uses of the building. Defects result from the progressive deterioration of the various components that make up a building. Defects occur through the action of one or a combination of external forces or agents. These have been classified in ISO 19208 [34] into five major groups as follows: mechanical agents (e.g. loads, thermal and moisture expansion, wind, vibration and noises); electro-magnetic agents (e.g. solar radiation, radioactive radiation, lightening, magnetic fields); thermal agents (e.g. heat, frost); chemical agents (water, solvents, oxidising agents, sulphides, acids, salts); and biological agents (e.g. vegetable, microbial, animal). Carillion [35] presents a brief description of the salient features of these five groups followed by a detailed description of numerous defects in buildings including symptoms, investigations, diagnosis and cure. Defects in buildings have been the subject of investigation over a number of decades and details can be found in many standard texts [36–39].

Dampness is increasingly significant as the value of a building can be affected even where only low levels of dampness are found. It is increasing seen as a health hazard in buildings. The fabric of buildings under normal conditions contains a surprising amount of moisture which does not cause any harm. The term dampness is commonly reserved for conditions under which moisture is present in sufficient quantity to become either perceptive to sight or by touch, or to cause deterioration in the decorations and eventually in the fabric of the building [42]. A building is considered to be damp only if the moisture becomes visible through discoloration and staining of finishes, or causes mould growth on surfaces, sulphate attack or frost damage, or even drips or puddles [42]. All of these signs confirm that other damage may be occurring.

There are various forms of decay which commonly afflict the fabric of buildings which can be attributed to the presence of excessive dampness. Dampness is thus, a catalyst for a whole variety of building defects. Corrosion of metals, fungal attack of timber, efflorescence of brick and mortar, sulphate attack of masonry materials, and carbonation of concrete are all triggered by moisture. Moisture in buildings is a problem because it expands and contracts with temperature fluctuations, causing degradation of materials over time. Moisture in buildings contains a variety of chemical elements and compounds carried up from the soil or from the fabric itself which can attack building materials physically or chemically. Damp conditions encourage the growth of wood rutting fungi, the infestation of timber by wood boring insects and the corrosion of metals. It enables the house dust mite population to multiply rapidly and moulds to grow, creating conditions which are uncomfortable or detrimental to the health of occupants.

*2.1 The Causes of Dampness*

A high proportion of dampness problems are caused by condensation, rain penetration, and rising damp [43]. Condensation occurs when warm air in the atmosphere reverts to water when it comes into contact with cold surfaces that are below the dew-point temperature. It is most prevalent in winter when activities such as cooking, showering and central heating release warm moisture into the air inside the building. The condensate can be absorbed by porous surfaces or appear as tiny droplets on hard shiny surfaces. Thus, condensation is dependent on the temperature of surfaces and the humidity of the surrounding air. Condensation is the most common damp that can be found in both commercial and residential property.

Rain penetration occurs mostly through roofs and walls exposed to the prevailing wind-driven rainfall, with openings that permit its passage and forces to drive or draw it inwards. It can be caused by the effects of incorrect design, bad workmanship, and structural movement, the wrong choice of or decay of material, badly executed repairs or lack of regular maintenance [44]. The most exposed parts of a building such as roofs, chimneys and parapets are the most susceptible to rain penetration.

Rising damp occurs through the absorption of water, by a physical process called capillary action, at the lower sections of walls and ground supported structures that are in contact with damp soil. The absorbed moisture rises to a height at which there is a balance between the rate of evaporation and the rate at which it can be drawn up by capillary action [44]. Rising damp is commonly found in older properties where the damp proof course is damaged or is absent. Other causes of dampness include: construction moisture; pipe leakage; leakage at roof features and abutments; spillage; ground and surface water; and contaminating salts in solution.

*2.2 The Effects of Damp*

Moisture can damage the building structure, the finishing and furnishing materials and can increase the heat transfer through the envelope and thus the overall building energy consumption [25]. It may also cause a poor indoor air quality and respiratory illness in occupants [42]. In the main, the types of deterioration driven by water in building materials include: moulds and fungal growth; materials spalling; blistering; shrinkage; cracking and crazing; irreversible expansion; embrittlement; strength loss; staining or discolouration; steel and iron rusting; and decay from micro-organisms. In extreme cases, mortar or plaster may fall away from the affected wall. The focus of this paper is on moulds, stains and paint deterioration which are the most common interrelated defects arising from dampness.

Moulds and fungi occur on interior and exterior surfaces of buildings following persistent condensation or other forms of frequent dampness. On internal surfaces, they are unsightly and can cause a variety of adverse health effects including respiratory problems in susceptible individuals [42]. Moulds on external surfaces are also unsightly and cause failure of paint films. Fungi growth such as algae, lichens and mosses generally occur on external surfaces where high moisture levels persist for long periods [45,46]. They are unsightly and can cause paint failure like moulds.

Paint deterioration typically occurs on building surfaces exposed to excessive moisture, dampness and other factors due to anthropogenic activity. The daily and seasonal variations in temperatures that oscillate between cold and warm, cause the moisture to expand and contract continuously resulting in paint slowly separating from a building's surface. The growth of mildew and fungi in the presence of moisture and humid conditions will cause paint damage. The exposure of exterior surfaces to sun and its ultra-violet radiation causes paint to fade and look dull. Exposure to airborne salts and pollution will also cause paint deterioration. Paint deterioration is visible in the form of peeling, blistering, flacking, blooming, chalking, and crazing.

## 3. Convolutional Neural Networks (ConvNet)

CNN, a class of deep learning techniques, are primarily used for solving fundamental problems in computer vision such as image classification [47], object detection [23,48], localisation [25] and segmentation [49]. Although early deep neural networks (DNN) go back to the 1980's when Fukushima [50] applied them for visual pattern recognition, they were not widely used, except in few applications, mainly due to limitation in the computational power of the hardware which is needed to train the network. It was in mid-2000s when the developments in computing power and the emergence of large amounts of labelled datasets contributed to deep learning advancement and brought CNN back to light [51].

*3.1 CNN Architecture*

The simplest form of a neural network is called perceptron. This is a single-layer neural network with exactly one input layer and one output layer. Multiple perceptrons can be connected together to form a multi-layer neural network with one input, one output and multiple inner layers, which also known as hidden layers. The more hidden layers, the deeper is the neural network (hence the name deep neural network) [52]. As a rule of thumb when designing a neural network, the number of nodes in the input layer is equal to the number of features in the input data, e.g. since our inputs are images with 3-channel (Red, Green, Blue) with 224x 224 pixels in each channel, therefore, the number of nodes in our input layer is 3x224x224. The number of nodes in the output layer, on the other hand, is determined by the configuration of the neural network. For example, if the neural network is a classifier, then the output layer needs one node per class label, e.g. in our neural network, we have four nodes corresponding to the four class labels: mould, stain, deterioration and normal.

When designing a neural network, there are no particular rules that govern the number of hidden layers needed for a particular task. One consensus on this matter is how the performance changes when adding additional hidden layers, i.e. the situations in which performance improves or becomes worse with a second (or third, etc.) hidden layer [53]. There are, however, some "empirical-driven" rules of thumbs about the number of nodes (the size) in each hidden layer [54]. One common way suggest that the optimal size of the hidden layer should be between the size of the input layer and the size of the output layer [55].

3.1.1 CNN Layers

Although, CNN have different architectures, almost all follow the same general design principles of successively applying convolutional layers and pooling layers to an input image. In such arrangement, the ConvNet continuously reduces the spatial dimensions of the input from previous layer while increasing the number of features extracted from the input image (see Figure 1).

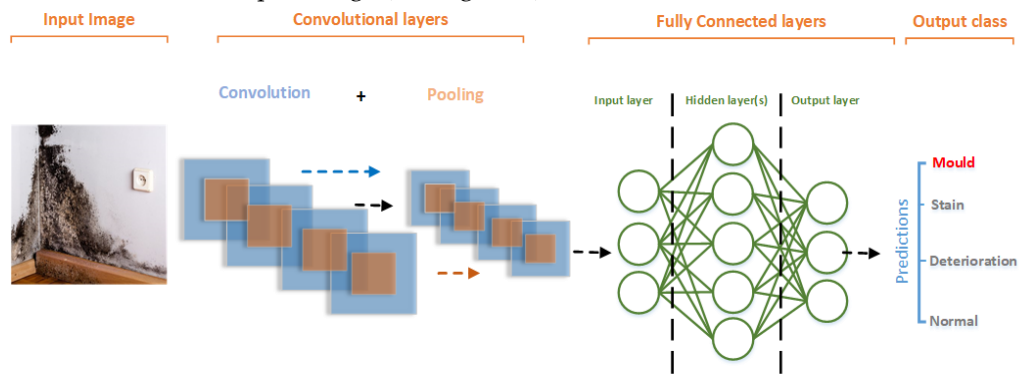

**Figure 1.** Basic ConvNet Architecture.

Input images in neural networks are expressed as multi-dimensional arrays where each colour pixel is represented by a number between 0 and 255. Grey scale images are represented by a 1-D array, while RGB images are represented by a 3-D array, where the colour channels (Red, Green and Blue) represent the depth of the array. In the convolutional layers, different filters with smaller dimensions arrays but same depth as the input image (dimensions can be 1x1xm, 3x3xm, or 5x5xm, where m is the depth of the input image), are used to detect the presence of specific features or patterns present in the original image. The filter slides (convolved) over the whole image starting at the top left corner while computing the dot product of the pixel value in the original image with the values in the filter to generate a feature map. ConvNets use pooling layers to reduce the spatial size of the network by breaking down the output of the convolution layers into smaller regions were the maximum value of every smaller region is taken out and the rest is dropped (max-pooling) or the average of all values is computed (average-pooling) [56–58]. As a result, the number of parameters and computation required in the neural network is reduced significantly.

The next series of layer in ConvNets are the fully connected (FC) layers. As the name suggests, it is a fully connected network of neurons (perceptrons). Every neuron in one sub-layer within the FC network, has a connection with all neurons in the successive sub-layer (see Figure 1).

At the output layer, a classification which is based on the features extracted by the previous layers is performed. Typically, for a multi-classifier neural network, a Softmax activation function is considered, which outputs a probability (a number ranging from 0-1) for each of the classification labels which the network is trying to predict.

*3.2 Transfer Learning*

According to a study by Zhu et al. [59], data dependence is one of the most challenging problems in deep learning where sufficient training requires huge amounts of information in order for the network to understand the patterns of data. In deep learning, both training and testing data are assumed to have same distribution and same feature space. In reality, however, adequate training data may exist in one domain, while a classification task is conducted on another. Moreover, if the data distribution changes on the target domain, a whole re-build of the classification network is required with a newly collected training dataset. In some applications, constructing largely-enough, properly-labelled datasets data can be quite challenging, particularly, when data acquisition is expensive or when data annotation is the timely consuming process. This could limit the development of many deep learning applications such as biomedical imaging where training data in most cases is not enough to train a network. Under these circumstances, transfer learning (knowledge) from one domain to another can be advantageous [32].

3.2.1 Mathematical Notations

Given a domain
$$\mathcal{D} = \{\mathcal{X}, P(X)\}, \tag{1}$$

Then $\mathcal{X}$ is the feature map, and $P(X)$ is the probability distribution, $X = \{x_1, \cdots, x_n\}$
and $\{x_1, \cdots, x_n\} \in \mathcal{X}$ then the learning task $\mathcal{T}$ is defined as
$$\mathcal{T} = \{\mathcal{Y}, \hat{y} = P(Y|)\} \tag{2}$$

Where $\mathcal{Y}$ is the set of all labels, and $\hat{y}$ is the prediction function. The training data is represented by the pairs $\{x_i, y_i\}$ where $x_i \in X, y_i \in \mathcal{Y}$.

Suppose $\mathcal{D}_S$ denotes the source domain and $\mathcal{D}_T$ denotes the target domain, then the source domain data can be presented as:

$$\mathcal{D}_S = \{(x_{S_1}, y_{S_1}), \cdots, (x_{S_n}, y_{S_n})\}, \tag{3}$$

where $x_{S_i} \in \mathcal{X}_S$ is the given input data, and $y_{S_i} \in \mathcal{Y}_S$ is the corresponding label. The target domain DT can also be represented in the same way:

$$\mathcal{D}_T = \{(x_{T_1}, y_{T_1}), \cdots, (x_{Tn}, y_{T_n})\}, \tag{4}$$

where $x_{T_i} \in \mathcal{X}_T$ is the given input data, and $y_{T_i} \in \mathcal{Y}_T$ is the corresponding label. In almost all real-life applications, the number of data instances in the target domain is significantly less than those in the source domain that is

$$0 \leq n_T \ll n_S \tag{5}$$

*Definition*

For a given source domain $\mathcal{D}_S$ and a learning task $\mathcal{T}_S$, with target domain $\mathcal{D}_T$ and a learning task $\mathcal{T}_T$, then transfer learning is the use of knowledge in $\mathcal{D}_S$ and $\mathcal{T}_S$ to improve the learning of the prediction function $\hat{y}_T$ in $\mathcal{D}_T$, given that $\mathcal{D}_S \neq \mathcal{D}_T$ and $\mathcal{T}_S \neq \mathcal{T}_T$ [60].

Since any domain is defined by the pair

$$\mathcal{D} = \{\mathcal{X}, P(X)\}, \tag{6}$$

where $\mathcal{X}$ is the feature map, and $P(X)$ is the probability distribution, $X = \{x_1, \cdots, x_n\} \in X$,

then according to the definition, if $\mathcal{D}_S \neq \mathcal{D}_T$

$$\Rightarrow \quad \mathcal{X}_S \neq \mathcal{X}_T \text{ and } P_S(X) \neq P_T(X). \tag{7}$$

Similarly, if a task is defined by

$$\mathcal{T} = \{\mathcal{Y}, \hat{y} = P(Y|X)\} \tag{8}$$

where $\mathcal{Y}$ is the set of all labels, and $\hat{y}$ is the prediction function, then by definition, if $\mathcal{T}_S \neq \mathcal{T}_T$

$$\Rightarrow \quad \mathcal{Y}_S \neq \mathcal{Y}_T \text{ and, } \hat{y}_S \neq \hat{y}_T. \tag{9}$$

Hence, the four possible scenarios of transfer learning are as follows:
1. When both target and source domains are different, that is when $\mathcal{D}_S \neq \mathcal{D}_T$ and their feature spaces are also different, i.e. $\mathcal{X}_S \neq \mathcal{X}_T$.

2. When the two domains are different, $\mathcal{D}_S \neq \mathcal{D}_T$ and their probability distribution are also different, i.e. $P_S(X) \neq P_T(X)$, where $x_{T_i} \in \mathcal{X}_T$, and $P_S(X) \neq P_T(X)$.
3. When the two domains are different, $\mathcal{D}_S \neq \mathcal{D}_T$ and both their learning tasks and label spaces are different, that is $\mathcal{T}_S \neq \mathcal{T}_T$S and, $\mathcal{Y}_S \neq \mathcal{Y}_T$, respectively.
4. When the target domain $\mathcal{D}_T$ and a source domain $\mathcal{D}_S$ are different, and their conditional probability distributions are also different, that is, when $\hat{y}_T \neq \hat{y}_S$,

where
$$\hat{y}_T = P(Y_T|X_T), \qquad (10)$$

and
$$\hat{y}_S = P(Y_S|X_S), \qquad (11)$$

such that
$$Y_{Ti} \in \mathcal{Y}_T, Y_{Si} \in \mathcal{Y}_S.$$

If both source and target domains are the same, that is when $\mathcal{D}_S = \mathcal{D}_T$, the learning tasks of both domains are also the same (i.e $\mathcal{T}_S = \mathcal{T}_T$). In this scenario, the learning problem of the target domain becomes a traditional machine learning approach and transfer learning is not necessary.

In ConvNets, transfer learning refers to using the weights of a well-trained network as an initialiser for a new network. In our research, we utilized the knowledge gained by a trained VGGNET on ImageNet [61] dataset which is a set that contains 14 million annotated images and contains more than 20,000 categories to classify images containing mould, stain and paint deterioration.

3.2.2 Fine-tuning

Fine-tuning is a way of applying transfer learning by taking a network that is already been trained for some given task and then tune (or tweaks) the architecture of this network to make it perform a similar task. By tweaking the architecture, we mean removing one or more layers of the original model and adding new layer(s) back to perform a new (similar) task. The number of nodes in the input and output layers in the original model also need to be adjusted in order to match the configuration of the new model. Once the architecture of the original model has bees modified, we then want to *freeze* the layers in the new model that came from the original model. By *freezing*, we mean that we want the weights for these layers unchanged when the new (modified) model is being re-trained on the new dataset for the new task. In this arrangement, only the weights of the new (or modified) layers are updating during the re-training and the weights of the *frozen* layers are kept the same as they were after being trained on the original task [62].

The amount of change in the original model, i.e. the number of layers to be replaced, primarily, depends on the size of the new dataset and its similarity to the original dataset (e.g. ImageNet-like in terms of the content of images and the classes, or very different, such as microscope images). When the two datasets have high similarities, replacing the last layer with a new one for performing the new task is sufficient. In this case, we say, transfer learning is applied as a classifier [63]. In some problems, one may want to remove more than just the last single layer, and add more than just one layer. In this case, we say, the modified model acts as feature extractor (see Figure 2) [60].

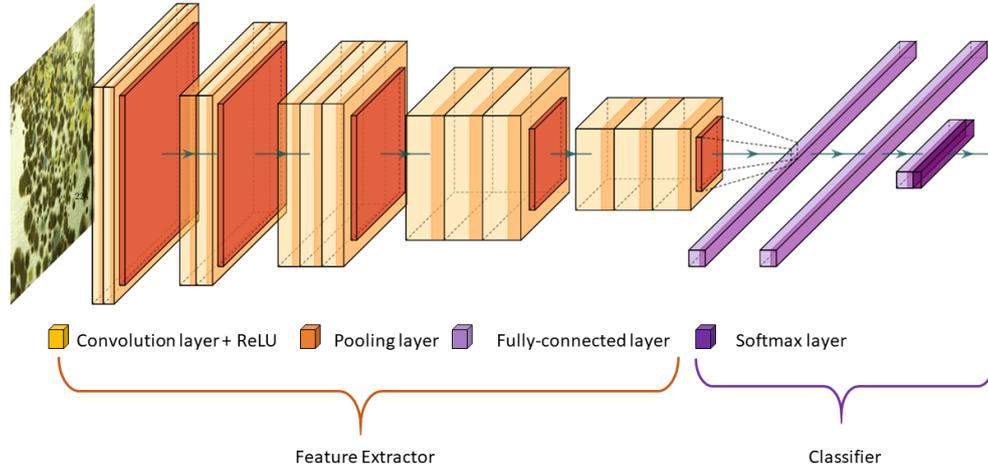

**Figure 2.** VGG-16 model. Illustration of using the VGG-16 for transfer learning. The convolution layers can be used as features extractor, and the fully connected layers can be trained as a classifier.

**3.4 Object Localisation Using Class Activation Mapping (CAM)**

The problem of object localisation is different from image classification problem. In the latter, when an algorithm looks at an image it is responsible for saying which class this image belongs to. For example, our model is responsible of saying this image is a "Mould", or a "Stain", or a "Paint deterioration" or "Normal". In the localisation problem however, the algorithm is not only responsible for determining the class of the image, it is also responsible for locating existing objects in any one image, and labelling them, usually by putting a rectangular bounding box to show the confidence of existence [64]. In the localisation problem, in addition to predicting the label of the image, the output of the neural network also returns four numbers (x0, y0, width, and height) which parameterise the bounding box of the detected object. This task requires different ConvNet architecture with additional blocks of networks called Regional Proposal Networks and Boundary-Box regression classifiers. The success of these methods however, rely heavily, on training datasets containing lots of accurately annotated images. A detailed image annotation, e.g. manually tracing an object or generating bounding boxes, however, is both expensive and often timely consuming [64].

A study by Zhou et al. [41] on the other hand, has shown that some layers in a ConvNet can behave as object detectors without the need to provide training on the location of the object. This unique ability, however, is lost when fully-connected layers are used for classification.

CAM is a computational-low-cost technique used with classification-trained ConvNets for identifying discriminative regions in an image. In other words, CAM highlights the regions in an image which are relevant to a specific class by re-using classifier layers in the ConvNet for obtaining good localisation results. It was first proposed by Zhou et al. [65] to enable classification-trained ConvNets to learn to perform object localisation without using any bounding box annotations. Instead, the technique allows the visualisation of the predicted class scores on an input image by highlighting the discriminative object parts which were detected by the neural network. In order to use CAM, the network architecture must be slightly modified by adding a global average pooling (GAP) after the final convolution layer. This new GAP is then used as a new feature map for the fully-connected layer which generates the required (classification) output. The weights of the output layer is then projected back on to the convolutional feature maps allowing a network to identify the importance of the image regions. Although simple, the CAM technique uses this connectivity structure to interpret the prediction decision made by the network. The CAM technique was applied to our model and was able to accurately, localise defects in images as depicted in Figure 9.

4. Methodology

The aim of this research is to develop a model that classifies defects arising from dampness as "mould", "stain" or "deterioration", should they appear in a given image, or as "normal" otherwise. In this work we also examine the extent of ConvNets role in addressing challenges arising from the nature of the defects under investigation and the surrounding environment. For example, according to one study, mould in houses can be black, brown, green, olive-green, gray, blue, white, yellow or even pink [66]. Moreover, stains and paint deterioration do not have a defined shape, size or colour and their physical characteristics are heavily influenced by the surrounding environment, i.e. the location (walls, ceilings, corners, etc.), the background(paint colour, wallpaper patterns, fabric, etc.) and by the intensity of light under which images of these defects were taken. The irregular nature of the defects imposes a big challenge when obtaining an adequate large-enough dataset to train a model to classify all these cases.

For the purpose of this research, images containing the defect types were collected from different sources. The images were then appropriately, cropped and resized to generate the dataset which was used to train our model. To achieve higher accuracy, instead of training a model from scratch, we adopted a transfer learning technique and used a pre-trained VGG-16 on ImageNet as our chosen model to customise and initialise weights. A separate set of images, not seen by the trained model, was used for validation to examine the robustness of our model. Finally, the CAM technique was applied to address the localisation problem. Details of the dataset, the tuned model and final results are discussed in the ensuing sections.

*4.1 Dataset*

Images of different resolutions and sizes were obtained from many resources, including photos taken by mobile phone, a hand-held camera, and copyright-free images obtained from the internet. These images were sliced into 224×224 thumbnails to increase the size of our dataset producing a total number of 2622 images in our dataset. The data was labelled into four main categories: normal (image containing no defects), mould, stain, and paint deterioration (which includes peeling, blistering, flacking, and crazing). The total number of images used as training data was 1890: mould (534 images), stain (449), paint deterioration (411), and normal (496). For the validation set, 20% of the training data (382 images out of the 1890 images) was randomly selected. In order to avoid overfitting and for better generalisation, a broad range of image augmentations were applied to the training set, including rescaling, rotation, height and width shift, horizontal and vertical flips. The remaining 732 images out of the 2622 were used as testing data with 183 images for each class. A sample of images used is shown in Figure 3

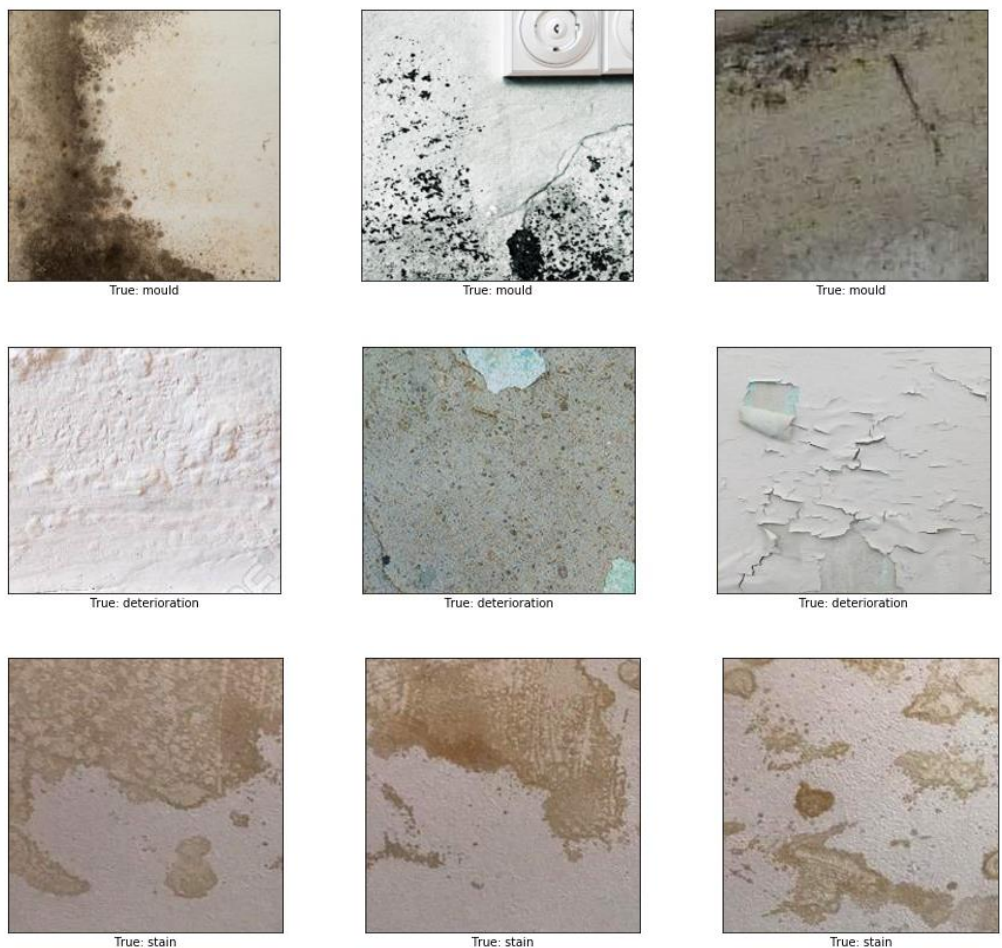

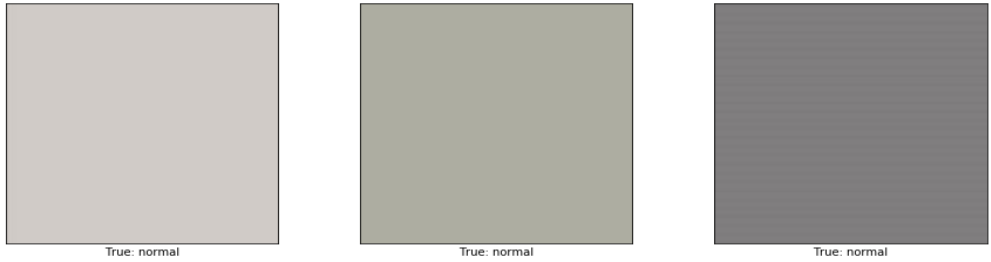

**Figure 3.** Dataset used in this study. A sample of the dataset that was used to train our model showing different mould images (first row), paint deterioration (second row), stains (third row).

*4.2 The model*

For our model, we applied a fine-tuning transfer learning to a VGG-16 network pre-trained on ImageNet; a huge dataset of images containing more than 14 million annotated images and more than 20,000 categories [60]. Our choice of using the VGG-16, is mainly because it is proven to be a powerful network although having a simple architecture. This simplicity makes it easier to modify for transfer learning and for the CAM technique without compromising the accuracy. Moreover, VGG-16 has fewer layers (shallower) than other models such as the ResNet50 or Inception. According to Kaiming et Al. [67], deeper neural networks are harder to train. Since the VGG-16 has fewer layers than other networks, it makes it a better model to train on our relatively small dataset compared to deeper neural networks as figure 5 shows, accuracy are close, however VGG-16 training is smoother.

The architecture of the VGG-16 model (illustrated in Figure 4) comprises five blocks of convolutional layers with max-pooling for feature extraction. The convolutional blocks are followed by three fully-connected layers and a final 1 X 1000 Softmax layer (classifier). The input to the ConvNet is a 3-channel (RGB) image with a fixed size of 224 × 224. The first block consists of two convolutional layers with 32 filters, each of size 3 × 3. The second, third, and forth convolution blocks use filters of sizes 64 × 64 × 3, 128 × 128 × 3, and 256 × 256 × 3 respectively.

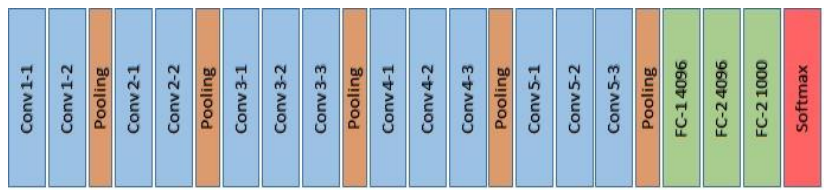

**Figure 4.** VGG-16 architecture. The VGG-16 model consists of 5 Convolution layers (in blue) each is followed by a pooling layer (in orange), and 3 fully-connected layers (in green), then a final Softmax classifier (in red).

For our model, we fine-tuned the VGG-16 model by, Firstly, *freezing* the early convolutional layers in the VGG-16, up to the fourth block, and used them as generic feature extractor. Secondly, we replaced the last 1 X 1000 Softmax layer (classifier) by a 1 X 4 classifier for classifying the 3 defects and a normal class. Finally, we re-trained the newly modified model allowing only the weights of block five to update during training.

Although different implementations of transfer learning were also examined (one by re-training the whole VGG-16 model on our dataset and replacing the last (Softmax) layer by 1 x 4 classifier, another implementation by freezing fewer layers and allowing weights of more layers to update during training), the arrangement mentioned earlier has proven to work better.

*4.3 Results*

**Class prediction**

The network was trained over 50 epochs using batch of size of 32 images and a step of 250 images per epoch. The final accuracy recorded at the end of the 50th epoch was 97.83% for training and 98.86% for the validation (Figure 5a). The final loss value was 0.0572 for training and 0.042 on the validation set (Figure 5b). The plot of accuracy in Figure 5.a, also shows that the model has trained well although the trend for accuracy on both validation, and training datasets is still rising for the last few epochs. It also shows that all models have not over-learned the training dataset, showing similar learning skills on both datasets despite the spiky nature of the validation curve. Similar learning pattern can also be observed from Figure 5b as both datasets are still converging for the last few epochs with a comparable performance on both training and validation datasets. Figure 5 also shows that all models had no overfitting problem during the training as the validation curve is converging adjacently to training curve and has not diverted away from the training curve.

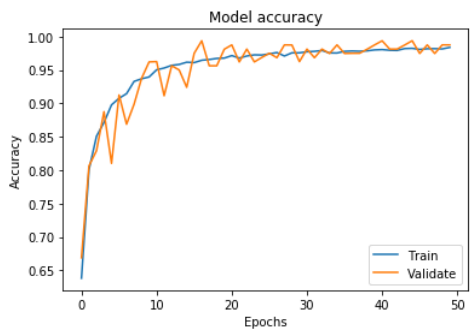 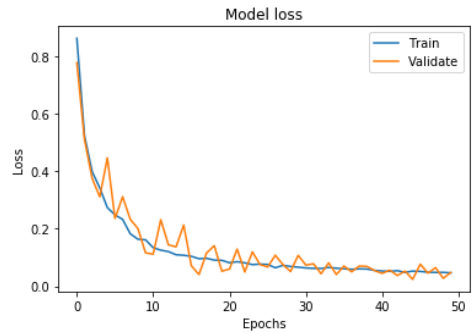

(a) VGG-16 Model accuracy        (b) VGG-16 Model loss

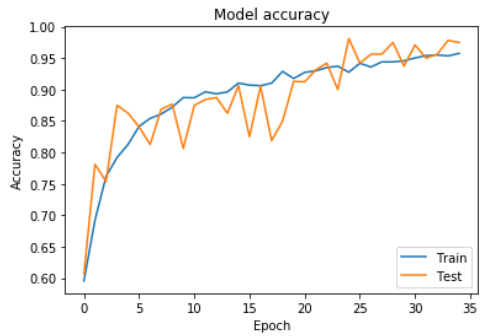
**(c) ResNet-50 model accuracy(96 / 95)**

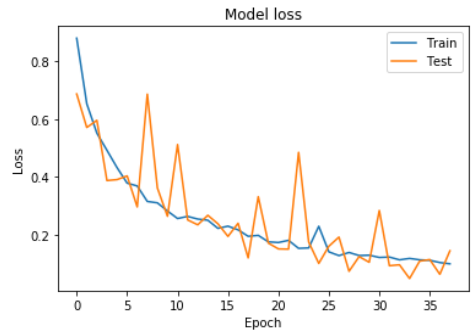
**(d) ResNet-50 model loss (0.103 / 0.103)**

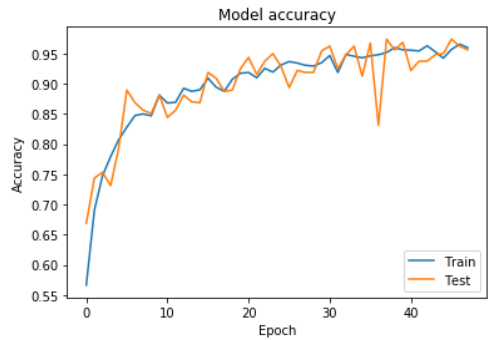
**(e) Inception model accuracy (96 / 95)**

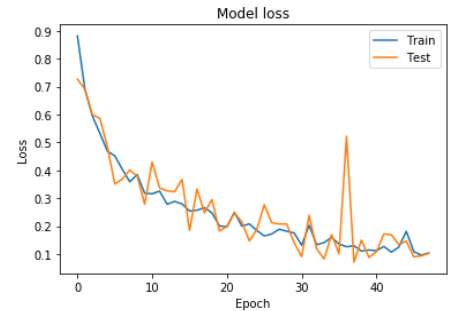
**(f) Inception model loss( 0.09 0.144)**

**Figure 5.** Comparative models accuracy and loss diagram. In sub-figure a) the VGG-16 model final accuracy is 97.83% for training and 98.86% for the validation. In sub-figure b) the final loss is 0.0572 for training and 0.042 on the validation set. In sub-figure c) the ReseNet-50 model final accuracy is 96.23% for training and 95.61% for the validation. In sub-figure d) the final loss is 0.103 for training and 0.102 on the validation set. In sub-figure e) the Inception model final accuracy is 96.77% for training and 95.42% for the validation. In sub-figure f) the final loss is 0.109 for training and 0.144 on the validation set.

To test the robustness of our model, we performed a prediction test on 732 non-used images dedicated for evaluating our model. The accuracy after completing the test was high and reached 87.50%. A sample example of correct classification of different types of defects is shown in Figure 6. In this figure, representative images show the accurate classification of our model to the four classes: in the first row, mould prediction (n= 167 out of 183), in the second row stain prediction, (n= 145 out of 183), in the third row deterioration prediction (n=157 out of 183) and in the fourth row the prediction of normal class (n= 183 out of 183). An example of miss-classified defects is shown in Figure 7. The representative images in this figure show failure of the model to correctly predict the correct class. 38 images containing stains were miss-classified; 31 as paint deterioration and 6 as mould and 1 as normal. 26 images containing paint deterioration were miss-classified; 13 as mould and 13 as stain. 16 images containing mould were miss-classified; 4 as paint deterioration and 12 as stain. The results in this figure illustrates an example where our model failed to identify the correct class of the damage caused by the damp. The false predictions for images by modern neural networks have been studied by many researcher [68–72]. According to Nguyen et Al. [68], although modern deep neural network achieved state-of-the-art performance and are able to classify objects in images with near-human-level accuracy, a slight change in an image, invisible to human eye can easily *fool* the neural network and cause it to miss-label the image. Guo et al. argues that this is primarily due to the fact that neural networks are overconfident in their predictions and often outputs highly confident predictions even with meaningless inputs [70]. In this work Guo et al studied the relationship between the accuracy of neural network and the predictions scores (confidence). According to the authors, a network with a confidence rate equal to accuracy rate is a *calibrated* neural network. However, they concluded that, although the capacity of neural networks has increased significantly in the past years, the increasing depth of modern neural networks negatively affects model calibration [70].

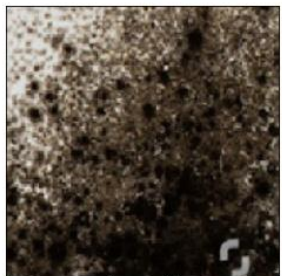 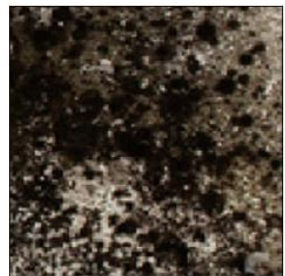 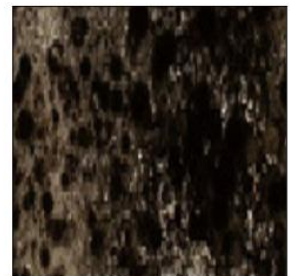
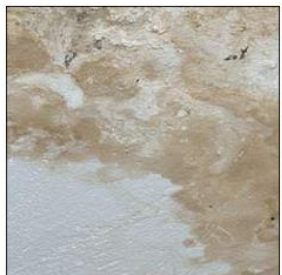 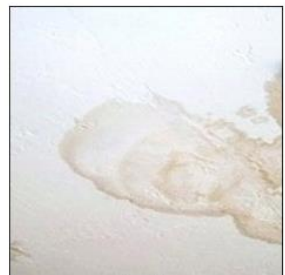 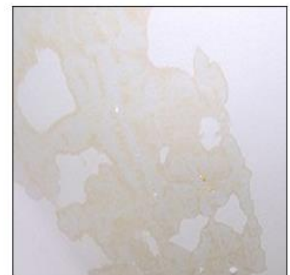
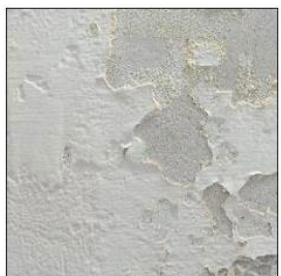 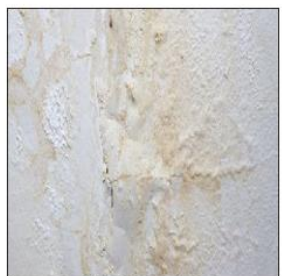 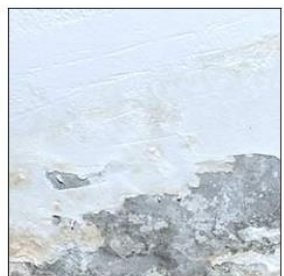

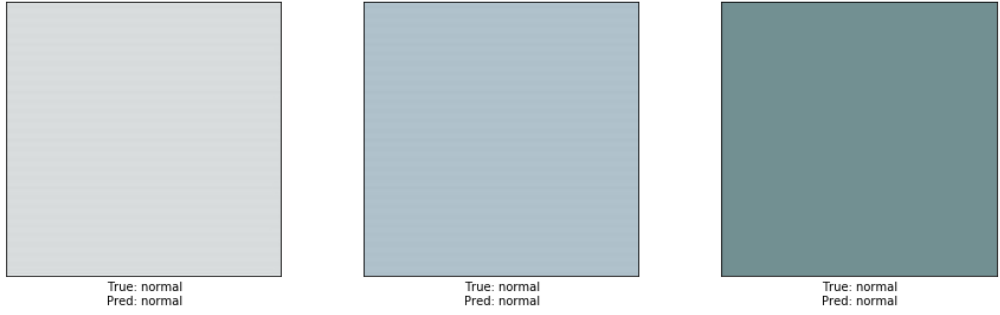

**Figure 6.** Correct classification of mould, stain and deterioration.

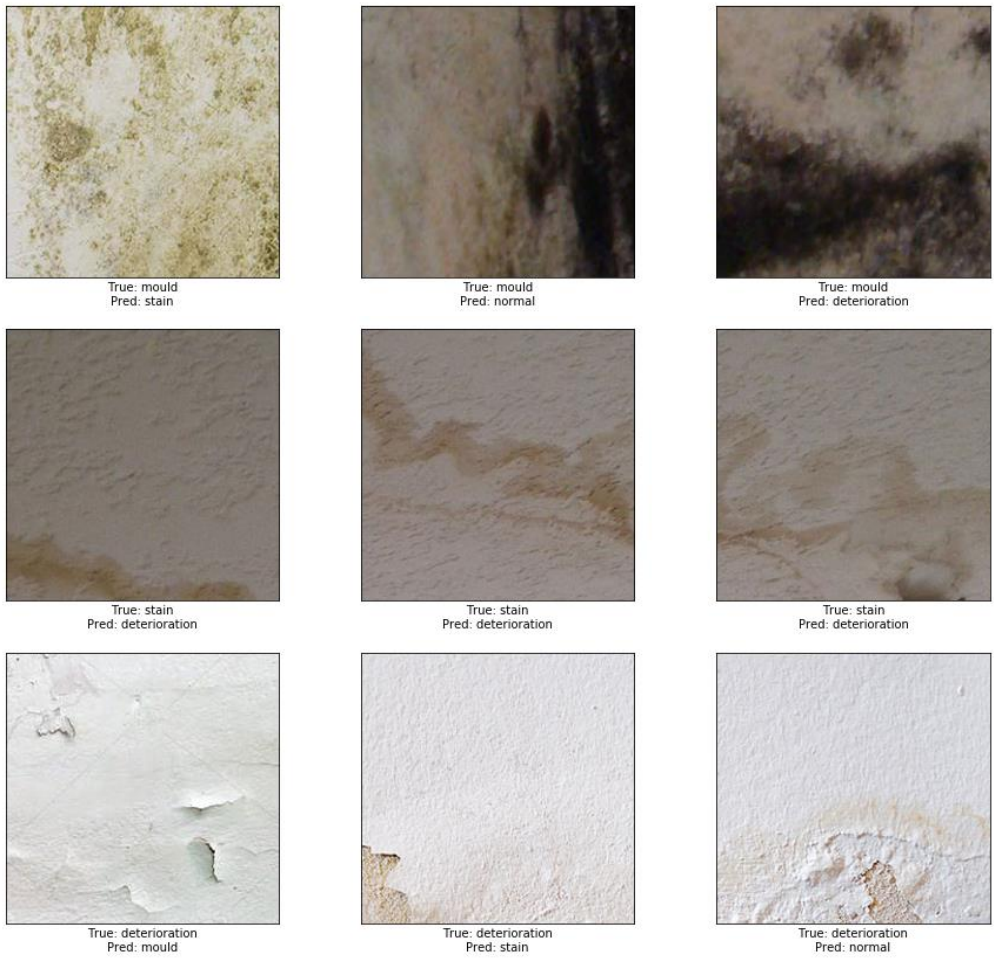

**Figure 7.** Example of miss-classification.

Our model, has performed very well on detecting images with mould with success rate of around 91%. The largest number of miss-classified defects occur in the stain and deterioration classes, with approximately 85% success rate in classifying stain and also 80% success rate in classifying paint deterioration (Figure 8). In Table 1, it can be seen that the overall precision of the model ranges between 82% for detecting deterioration, 84% for mould, and 89% for stain. The recall analysis show similar results, with 82% for detecting deterioration, 90% for mould, 99% for normal, and 89% for stain. Note that the precision which is the ability of the classifier to label positive classes as positive is the ratio tp / (tp + fp) where tp is the number of true positives and fp the number of false positives. The recall quantifies the ability of the classifier to find all the positive samples, that is the ratio tp / (tp + fn) where fn is number of false negatives. F1 Score is the weighted average of precision and recall, that is F1 Score = 2*(Recall * Precision) / (Recall + Precision).

**Defect localisation using CAM**

Following the high accuracy in detecting the three classes we then, asked the question: can our model detect the actual localisation of these classes? To do so, we integrated the CAM [41] with our network. The CAM is a technique that uses the gradient of an object under consideration entering the final convolutional layer of the ConvNet to produces a coarse localisation map which highlights the most significant regions in the image for predicting the class of this image. It can be seen from the images in Figure 10 with representative images showing different defects accurately localised using CAM method: in first row sample of images containing paint deterioration, in second row a sample of images containing stains, and in third and fourth rows samples of images containing mould that, our model has accurately allocated the defect in each image with high precision. In few cases, however, incorrect localisation was obtained, for example the images in Figure 11 showing some defect incorrectly located by our model. Although the first image contains paint deterioration spread over large areas of the image and the second image contains a large stain, both defects were miss-localised by our model. Figure 9 represents the schematic illustration of CNN-CAM localisation.

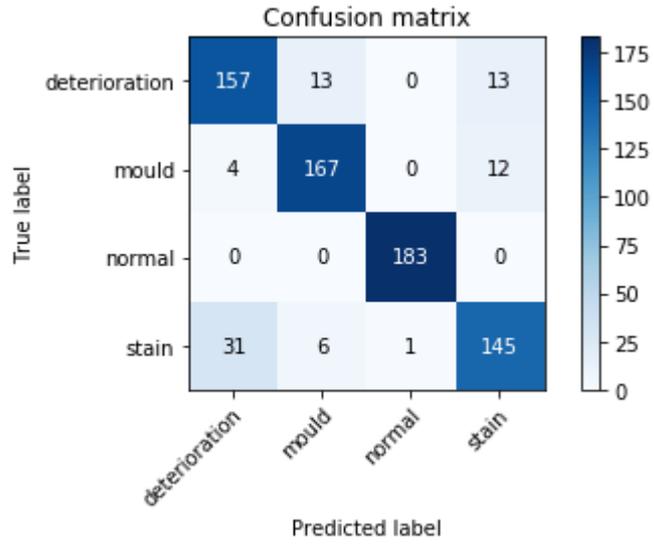

**Figure 8.** Confusion Matrix.

**Table 1.** Classification Report.

|  | Precision | Recall | F1-score | Support |
|---:|:---:|:---:|:---:|:---:|
| Deterioration | 0.82 | 0.86 | 0.84 | 183 |
| Mould | 0.90 | 0.91 | 0.91 | 183 |
| Normal | 0.99 | 1.00 | 1.00 | 183 |
| Stain | 0.89 | 0.79 | 0.82 | 183 |

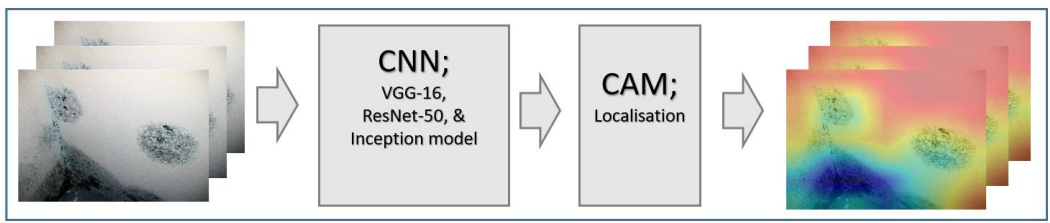

**Figure 9.** CNN-CAM localisation

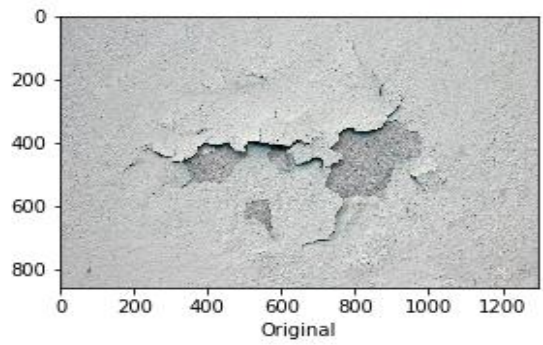
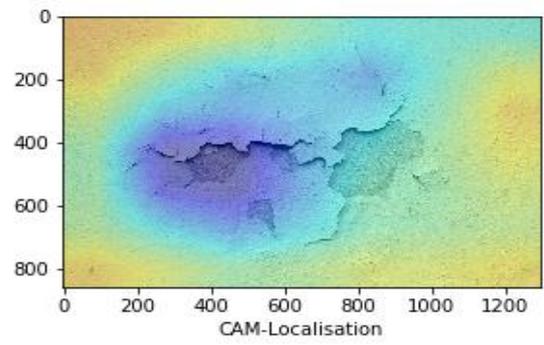
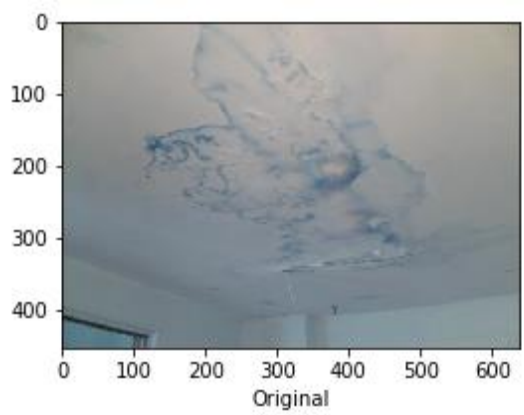
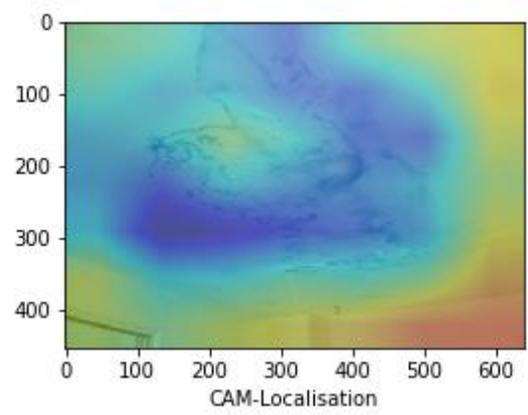
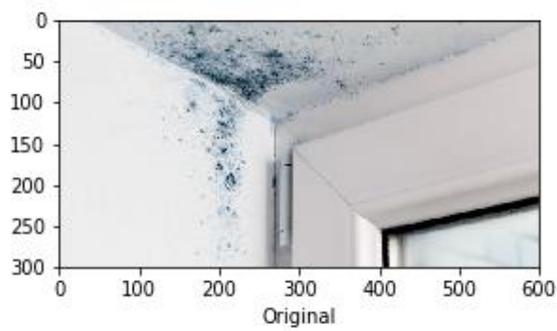
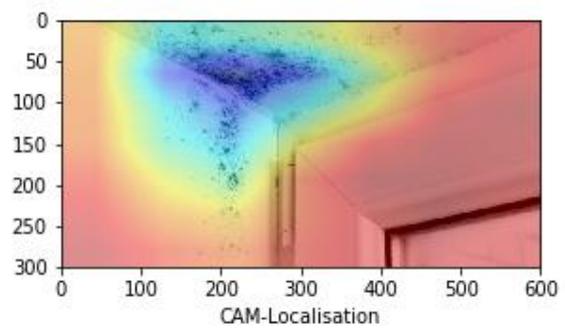

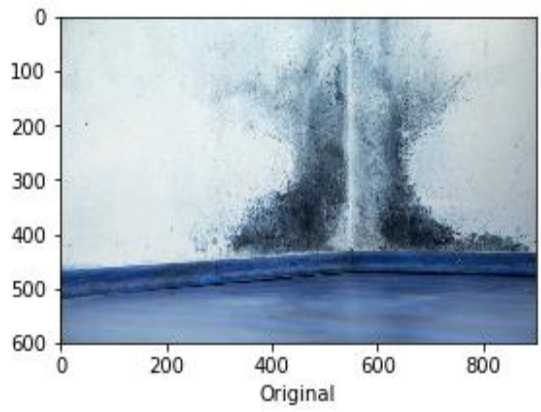
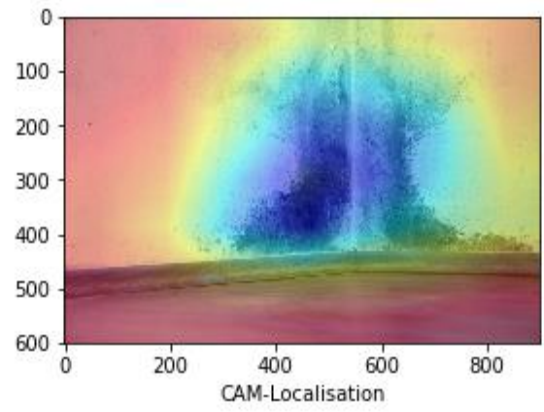

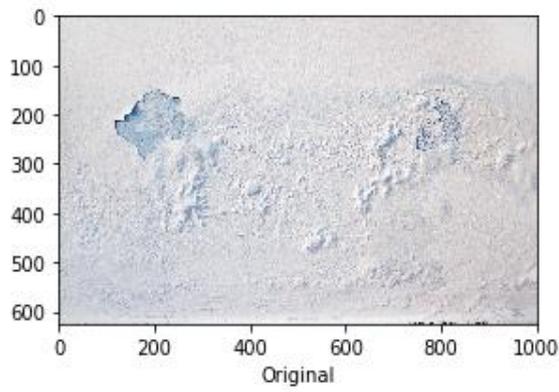
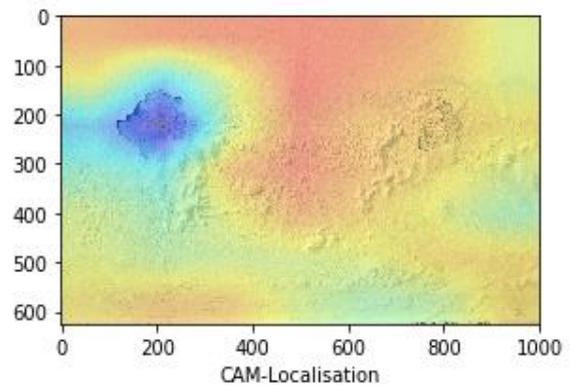

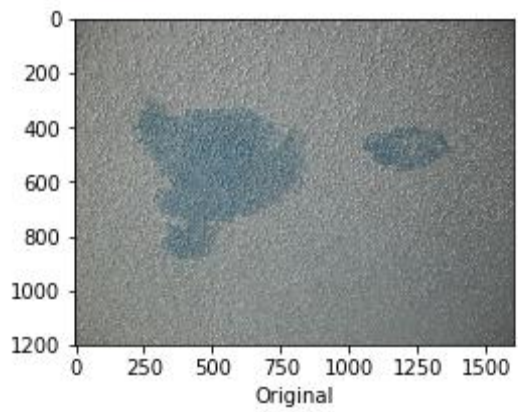
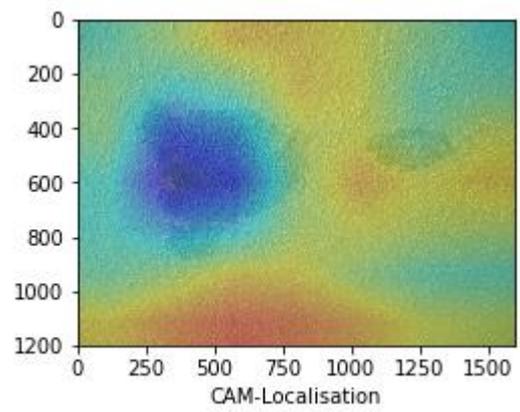

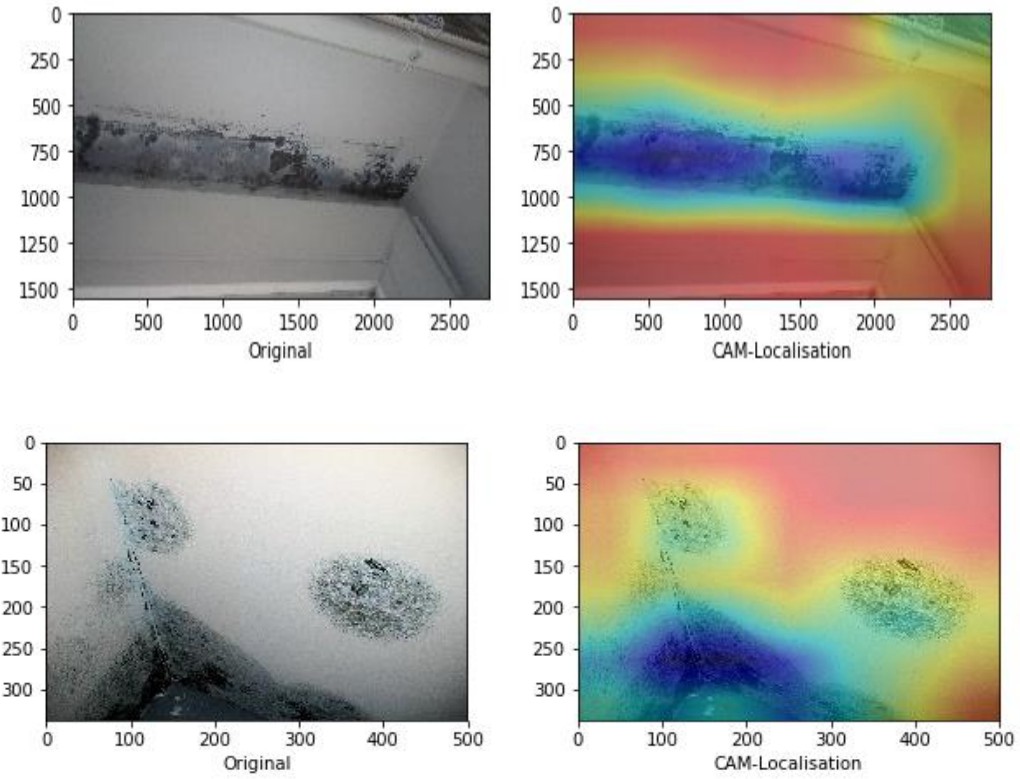

**Figure 10.** Correct localisation.

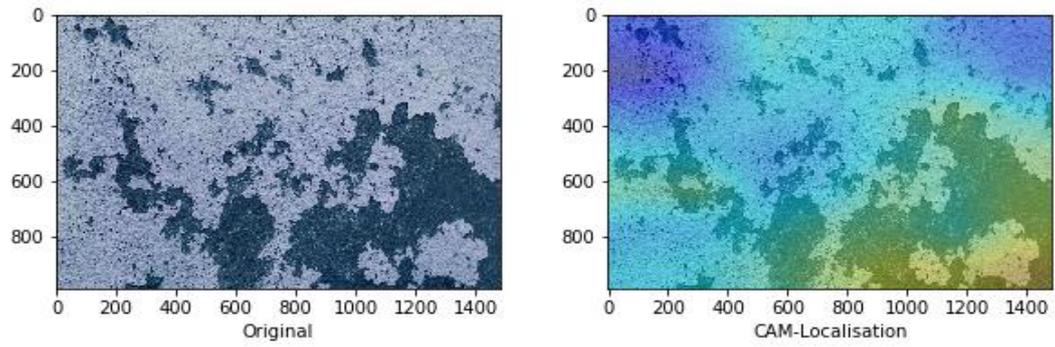

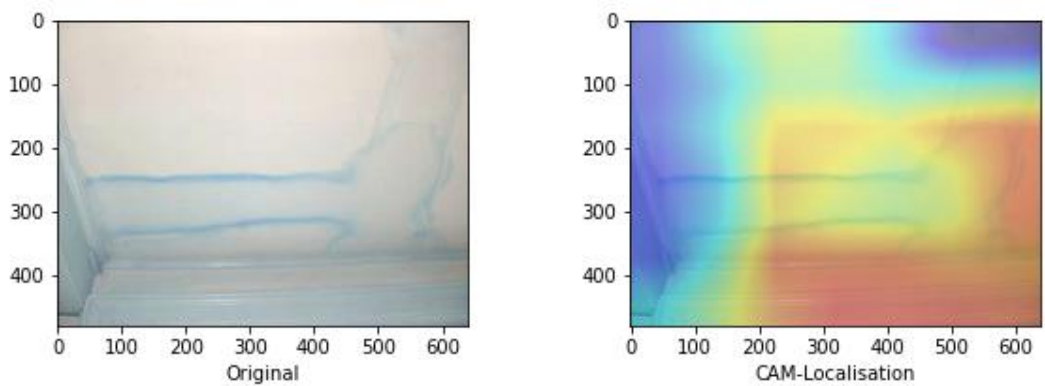

**Figure 11.** Incorrect localisation.

## 5. Discussion and Conclusions

The work is concerned with the development of a deep learning-based method for the automated detection and localisation of key building defects from given images. This research is part of work on condition assessment of built assets. The developed approach involves classification of images into four categories: three categories of building defects caused by dampness namely: mould, stain and paint deterioration which includes peeling, blistering, flacking, and crazing and of these defects and a fourth category "Normal" when no defects is present.

For our classification problem, we applied a fine-tuning transfer learning to a VGG-16 network pre-trained on ImageNet. A total of 2622 224x 224 images were used as our dataset. Out of the 2622 images, a total 1890 images used for training data: mould (534 images), stain (449), paint deterioration (411) and normal (496). In order to obtain sufficient robustness, we applied different augmentation techniques to generate larger dataset. For the validation set, a 20% of the training data (382 images) was randomly chosen. After 50 epochs, the network recorded an accuracy of 97.83% with 0.0572 loss on the training set and 98.86% with 0.042 loss on the validation. The robustness of our network was evaluated on a separate set of 732 images, 183 images for each class. The evaluation test showed a consistent overall accuracy of 87.50% and %90 of images containing mould correctly classified, 82% for images containing deterioration, 89% for images containing stain and 99% for normal images. To address the localisation problem, we integrated the CAM technique which was able to locate defects with high precision. The overall performance of the proposed network has shown high reliability and robustness in classifying and localising defects. The main challenge during this work was the availability of large labelled datasets which can be used to train a network for this type of problem. To overcome this obstacle, we used image augmentation techniques to generate synthetic data for a largely enough dataset to train our model.

The benefit of our approach, lays in the fact that, whilst similar research such as the one by Cha et al.[48] and others [26,28,59,73], focus only on detecting cracks on concrete surfaces which is a simple binary classification problem, we offer a method to build a powerful model that can accurately detect and classify multi-class defects given a relatively very small datasets. Whilst cracks in constructions have been widely studies and supported by large dedicated datasets such as [74] and [75], our work, describe herewith in, offers a roadmap for researcher in condition assessment of built assets to study other types of equally important defects which have not been addressed sufficiently.

For the future works, the challenges and limitation that we were facing this in paper will be addressed. The presented paper had to set a number of limitations, i.e., firstly, multiple types of the defects are not considered at once. This means that the images considered by the model belonged to only one category. Secondly, only the images with visible defects are considered. Thirdly, consideration of the extreme lighting and orientation, e.g., low lighting, too bright images, are not included in this study. In the future works, these limitations will be considered to be able to get closer to the concept of a fully automated detection. Through fully satisfying these challenges and limitations, our present work will be evolved into a software application to perform real-time detection of defects using vision sensors including drones. The work will also be extended to cover other models that can detect other defects in construction such as cracks, structural movements, spalling and corrosion. Our long-term vision includes plans to create a large, open source database of different building and construction defects which will support world-wide research on condition assessment of built assets.